\documentclass{article}

\usepackage[preprint]{neurips_2025}
\usepackage[utf8]{inputenc} % allow utf-8 input
\usepackage[T1]{fontenc}    % use 8-bit T1 fonts
\usepackage{url}            % simple URL typesetting
\usepackage{booktabs}       % professional-quality tables
\usepackage{amsfonts}       % blackboard math symbols
\usepackage{nicefrac}       % compact symbols for 1/2, etc.
\usepackage{microtype}      % microtypography
\PassOptionsToPackage{options}{natbib}
\usepackage{graphicx} 
\usepackage{multirow}
\usepackage{subcaption} 
\usepackage{overpic}
\usepackage[table]{xcolor}
\usepackage{colortbl} % 可选，增强颜色表格功能
\usepackage{fontawesome}

\def\aka{\emph{a.k.a.,~}}

\def\eg{\emph{e.g.,~}}
\def\etc{{\em etc}}

\newcommand*{\method}{$\mathbf{\mathcal{R}}$Bench-V}

\newcommand*{\sysone}{$\textbf{system-}\textbf{\uppercase\expandafter{\romannumeral 1}}$}
\newcommand*{\systwo}{$\textbf{system-}\textbf{\uppercase\expandafter{\romannumeral 2}}$}

\newcommand{\redline}[1]{\textcolor{red}{#1}}

\newcommand{\blue}[1]{\textcolor{blue}{#1}}
\newcommand{\red}[1]{\textcolor{red}{#1}}

\title{{\raisebox{-0.2\height}{\includegraphics[width=0.04\textwidth]{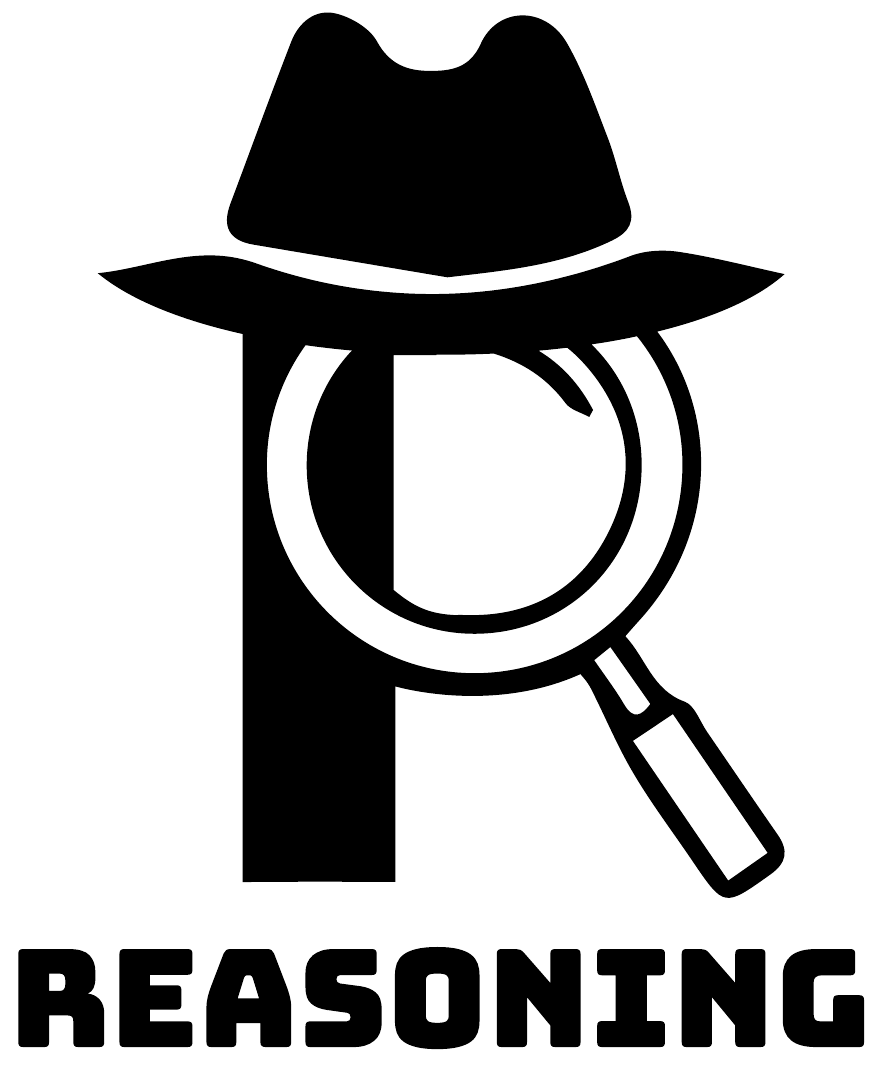}}}Bench-V: A Primary Assessment for \\
Visual Reasoning Models with Multi-modal Outputs}

\author{%
  Meng-Hao Guo$^{1}$, Xuanyu Chu$^{1}$, Qianrui Yang$^{1}$, Zhe-Han Mo$^{1}$,
  Yiqing Shen$^{1}$ \And Pei-Lin Li$^{1}$, 
  Xinjie Lin$^{1}$, Jinnian Zhang$^{2}$,
Xin-Sheng Chen$^{1}$, Yi Zhang$^{1}$ \And
Kiyohiro Nakayama$^{3}$, Zhengyang Geng$^{4}$, 
Houwen Peng$^{2}$, Han Hu$^{2}$,
Shi-Min Hu$^{1}$   
\thanks{Shi-Min Hu is the corresponding author. E-mail: shimin@tsinghua.edu.cn.}
\\
\\
$^{1}$ Tsinghua University, $^{2}$ Tencent Hunyuan X,   $^{3}$ Stanford University,   $^{4}$ Carnegie Mellon University  
}

\begin{document}

\maketitle

\begin{abstract}

The rapid advancement of native multi-modal models and omni-models, exemplified by GPT-4o, Gemini and o3 
with their capability to 
\textbf{process and generate}
content across modalities such as text and images,
marks a significant milestone in the evolution of intelligence.
% As these models increasingly underpin real-world applications, 
Systematic evaluation of their multi-modal output capabilities 
in visual thinking process
(\aka multi-modal chain of thought, M-CoT) 
becomes critically important.
However, existing benchmarks for evaluating multi-modal models primarily focus on assessing multi-modal inputs and text-only reasoning process
while neglecting the importance of reasoning through multi-modal outputs.
In this paper, 
we present a benchmark, dubbed as \method{}, designed to assess models’ vision-indispensable reasoning.
To conduct \method{}, we carefully hand-pick
803 questions covering math, physics, counting and games.
Unlike problems in previous benchmarks, which typically specify certain input modalities, 
\method{} presents problems centered on multi-modal outputs,
which require image manipulation, such as 
generating novel images and constructing auxiliary lines to support reasoning process.
% (e.g., requiring both images and text as input), 
% \method{} introduces problems that require visual reasoning, which means generating or modifying images during the problem-solving process, such as creating new images and constructing auxiliary lines.
We evaluate numerous open- and closed-source models on \method{}, including o3, Gemini 2.5 pro, Qwen2.5-VL, etc.
Even the best-performing model, o3, achieves only 25.8\% accuracy on \method{}, far below the human score of 82.3\%,
which shows current models struggle to leverage multi-modal reasoning.
% Besides, a case analysis is conducted, providing insights and directions for advancing model development.
Data and code are available at \url{https://evalmodels.github.io/rbenchv}.

\begin{figure}[t]
\begin{center}
\includegraphics[width=1.0\columnwidth]{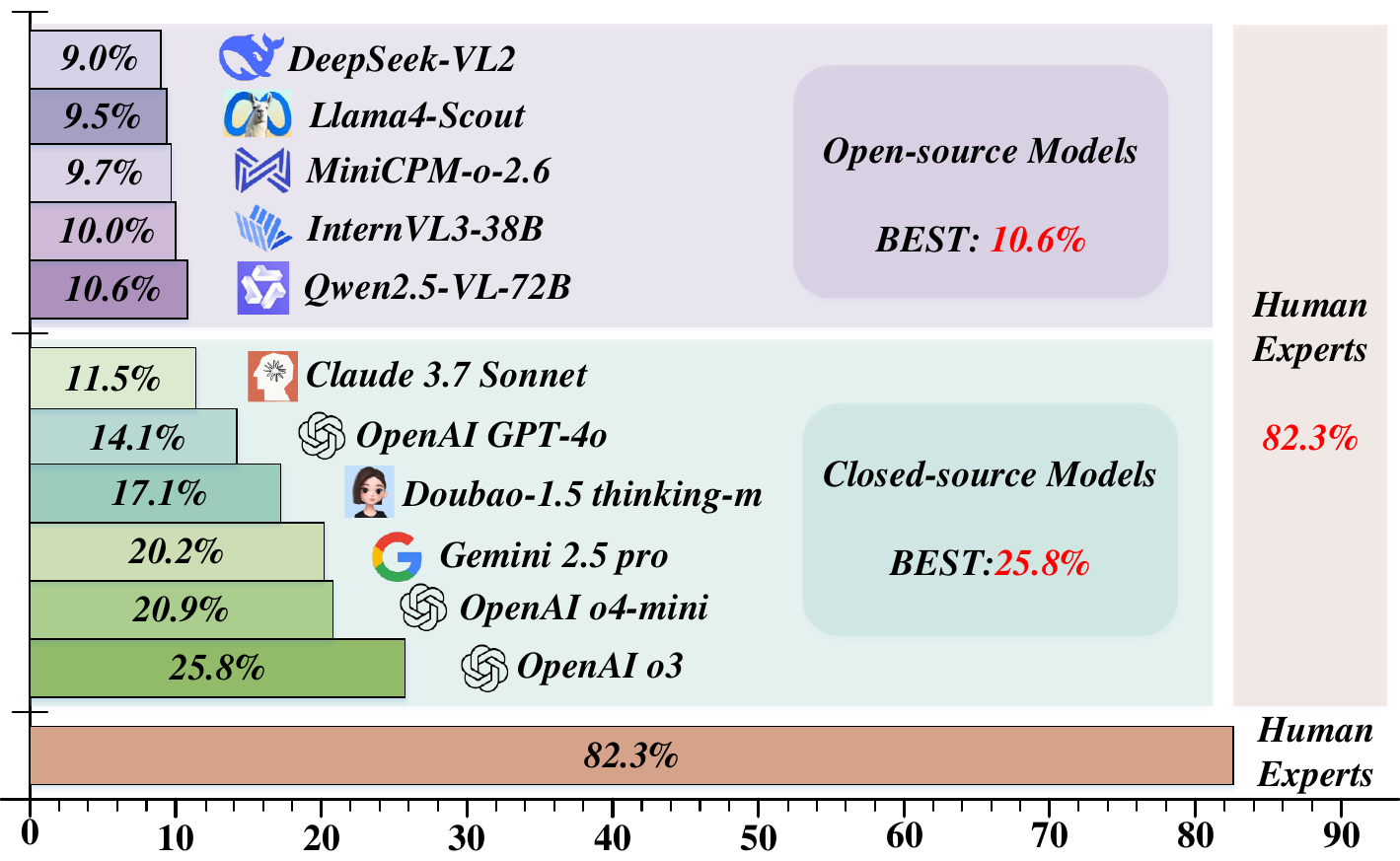}
\caption{The comparison between open-source models, closed-source models and human experts on \method{}.
It reveals there remains a significant gap between models and human experts in visual reasoning with multi-modal outputs.
}
\label{fig_comparsion_company}
\end{center}
\vskip -0.2in
\end{figure}

\end{abstract}

\section{Introduction}

\begin{flushleft}
\textit{“What I cannot create, I do not understand.”} 
 \quad — Richard Feynman.
\label{quote}
\end{flushleft}

Whether adults or children, when faced with complex problems,
they sometimes turn to drawing or diagramming to organize their thoughts, support reasoning, and seek solutions.
As highlighted by the quote~\ref{quote}  and findings in neuroscience~\cite{goldschmidt1991dialectics,pylyshyn2001visual,edwards2012drawing,fan2023drawing},  
the capability to draw during problem-solving is not only a hallmark of cognitive development but also an expression of human intelligence.
But what about intelligent models?
% As shown in Fig.~\ref{fig_motivation} (a),
Can they also learn to draw in order to reason and solve problems?

% Recently, 
% the landscape of foundation model has undergone a notable shift and  present a trend toward the two convergence: (1) modal convergence: from language models such as ChatGPT~\cite{chatgpt} to omni models like GPT-4o~\cite{gpt4oreport}; 
% (2) cognitive convergence: from chat models to reasoning models, which can be observed from ChatGPT~\cite{chatgpt}  to OpenAI o1~\cite{o1report}, and DeepSeek R1~\cite{guo2025deepseekr1}.
%Recently, the landscape of foundation models has undergone a profound  transformation, characterized by two major converging trends. 
Recently, researchers have made great progress toward equipping foundation models with above capability,  and the landscape of foundation models has undergone a profound transformation, driven by two major converging trends.
% the landscape of foundation models has undergone a profound  transformation, driven by two major converging trends. 
First, \textit{modal convergence} reflects the evolution from single-modality language models, such as ChatGPT~\cite{chatgpt}, to omni-modal systems capable of both multi-modal input and output, exemplified by GPT-4o~\cite{gpt4oreport}. 
Second, \textit{cognitive convergence} captures the transition from chat-oriented models to reasoning-driven models, as evidenced by the progression from ChatGPT~\cite{chatgpt} to more advanced systems such as OpenAI o1/o3~\cite{o1report} and DeepSeek R1~\cite{guo2025deepseekr1}.

As model input and output modalities converge, the evaluation frameworks for these leading foundation models must also evolve accordingly. Existing benchmarks such as MMMU~\cite{yue2024mmmu} and MMLU~\cite{hendrycks2021measuring},  have played a key role in advancing the field by providing frameworks to evaluate model capabilities.
However, these benchmarks are predominantly input-oriented, 
focusing on the model’s ability to interpret, understand, and reason over multi-modal inputs,
% evaluating the model's capacity to interpret and reason over multi-modal inputs using text-only outputs,
% such as images and text, 
while overlooking an equally critical aspect: the modality of outputs.
This refers to the model's ability to generate contextually appropriate multi-modal responses during the problem-solving process, whether through language, 
visual content, or other formats.

%As models advance in their ability to handle complex multi-modal scenarios, the ability to generate coherent and contextually appropriate multi-modal outputs (\emph{e.g.}, visual and textual outputs) for reasoning becomes increasingly critical.

%As ORMs continue to evolve rapidly, the design of evaluation benchmarks has become increasingly critical to ensure that these models are properly assessed for their performance across a variety of tasks and domains.

%Existing benchmarks such as MMMU~\cite{yue2024mmmu}, MMLU~\cite{hendrycks2021measuring}, and SEED~\cite{li2023seed} have played a significant role in advancing the field by providing frameworks to evaluate model capabilities.
%However, these benchmarks are predominantly input-oriented, focusing on the model’s ability to interpret, understand, and reason with multi-modal inputs,
% such as images and text, 
%while overlooking an equally important aspect: the modality of the output.
%It refers to the model's capacity to generate appropriate multi-modal responses during the problem-solving process, whether in language, visual content, or other formats. As models become more capable of handling complex, multi-modal scenarios, the ability to generate coherent and appropriate multi-modal outputs for reasoning is becoming increasingly crucial.

In this paper, we present \method{}, an early exploration for multi-modal output-oriented reasoning benchmark.
% Here, the letter "V" carries a dual meaning, signifying both the evaluation of \textbf{"V"}iusal thinking capabilities and the assessment of generating \textbf{"V"}isual content.
To build \method{}, 
we carefully and strictly
hand-pick and design 803 question-answer pairs, covering math, physics, counting, and games.
% Like examples shown in~\ref{fig_comparsion}, 
% We 
In Fig~\ref{fig_comparsion}, we clearly illustrate 
the differences between \method{} and 
other classic language model benchmarks, 
MMLU~\cite{hendrycks2021measuring}, and the multi-modal input-oriented benchmark, MMMU~\cite{yue2024mmmu}.
It can be seen that the main difference from previous benchmarks is that in \method{}, each question requires the model to produce multi-modal outputs during the reasoning process,
particularly modifications on the images, such as drawing images, adding auxiliary lines, and so on.

We evaluate a wide range of open- and closed-source 
multi-modal large language models (MLLMs) and omni models on the \method{}, including GPT-4o~\cite{gpt4oreport}, 
Gemini ~\cite{team2023gemini}, Qwen2.5VL~\cite{bai2025qwen2}, Claude3.5~\cite{claude35}, DeepSeek-VL2~\cite{wu2024deepseek}, \etc.
Besides, we also organize human to conduct tests on \method{}.
Our main observations and findings from the experiments are highlighted as follows. For more details, please refer to the experiment section.

\begin{figure}[t]
\begin{center}
\begin{overpic}[width=0.99\linewidth]{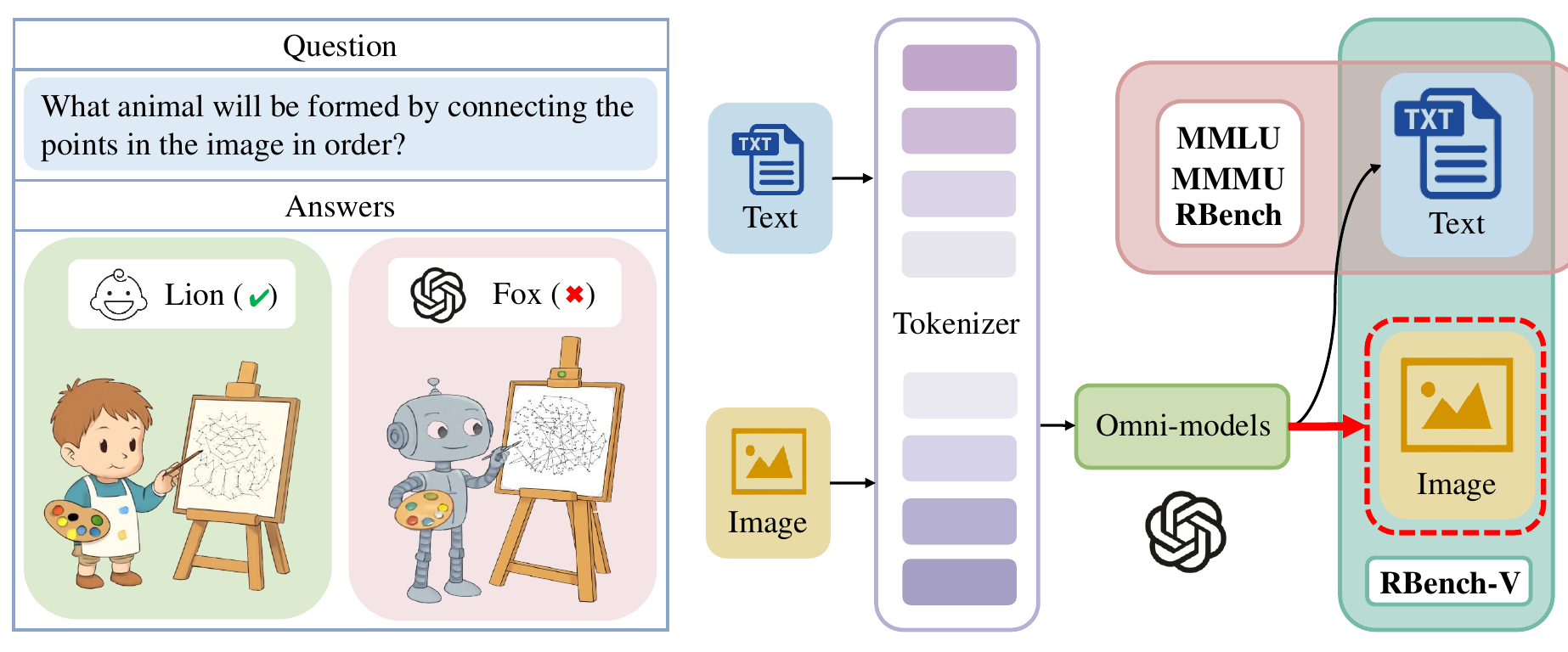}
  % \put(35, 0){Motivation of \method{}}
  % \put(65, 0){(b) Statistics of \method{}}
\end{overpic}\hspace{1pt}
\caption{The motivation of \method{}. Left: An illustration showing both humans and the GPT-4o model being asked a game-related question from \method{}.  
Right: This part shows common benchmarks such as MMLU, MMMU, and Rench focus on multi-modal inputs and textual outputs, whereas \method{} emphasizes not only multi-modal inputs but also multi-modal outputs.}
\label{fig_motivation}
\end{center}
\vskip -0.2in
\end{figure}

\begin{itemize}
    \item  If models, such as the InternVL or Qwen-VL series, lack multi-modal CoT, merely increasing their model sizes will not effectively resolve the challenges of vision-indisperential reasoning. %In both the Qwen series and the InternVL series of models, we observe that simply increasing model size does not appear to effectively address the challenge of multi-modal reasoning process.  
    It might be necessary to explore new paradigms, potentially incorporating M-CoT or agent-based reasoning frameworks, to solve visual-necessary complex problems.
    %To overcome this challenge, it may be necessary to explore new paradigms, potentially involving M-CoT or agent-based reasoning frameworks.
    \item  Despite the diverse capabilities of these models, even the highest-performing one, \emph{i.e.}, o3~\cite{gpto3o4}, achieves only 25.8\% accuracy on \method{}, which is significantly lower than the human score of 82.3\%. 
    This stark contrast highlights that, while impressive in many areas, the models still struggle to generate and integrate appropriate multi-modal responses in the visual thinking process.
    Besides, the o3 model has achieved significant progress in visual reasoning, outperforming previous state-of-the-art models by a substantial margin (+4.9\%). \method{} effectively captures this advancement and offers an automated framework for evaluating multi-modal output capabilities in visual reasoning.
    \item  The reasoning thought of models differ from that of human experts in math. We find that while models perform well on math questions, this does not necessarily indicate that they have learned to multi-modal reasoning. Instead, models often convert certain geometry problems into algebraic ones and solving them through textual reasoning. In contrast, human experts tend to prefer geometric solutions. This highlights a fundamental difference between the intelligence exhibited by current models and that of humans.
\end{itemize}

\section{Related Work}

\subsection{Foundation Models}
\label{sec_foundation_model}
% Since the emergence of ChatGPT~\cite{chatgpt}, 
Foundation models have evolved rapidly along two axes:
the expansion from language understanding to omni-models, and the progression from chat models to advanced reasoning models.

\paragraph{From language models to omni-models:} It represents progress along the modality axis. 
Early foundation models—such as ChatGPT~\cite{chatgpt}, LLaMA~\cite{touvron2023llama}, Qwen~\cite{bai2023qwen}, 
Mistral~\cite{jiang2023mistral},
GLM~\cite{zeng2022glm}, \etc, 
are limited to text-based dialogue. Researchers then begin exploring models with multi-modal inputs, including GPT-4V~\cite{gpt4vreport}, 
LLaVA~\cite{liu2024visual}, miniGPT-4~\cite{zhu2023minigpt}, Claude3.5~\cite{claude35}, \etc.
Recently, attention has shifted toward omni-models, which not only receive multi-modal inputs but also generate flexible multi-modal outputs (e.g., GPT-4o~\cite{gpt4oreport}, Gemini 2.5 Pro~\cite{gemini25pro}, Qwen2.5-Omni~\cite{xu2025qwen2} and Emu3~\cite{wang2024emu3}).

\paragraph{From chat models to reasoning models:}  
Another critical axis of advancement lies in reasoning capabilities.
Early chat-based foundation models~\cite{chatgpt,touvron2023llama,abdin2024phi} primarily focus on fluent and context-aware dialogue generation.
Recently, researchers have begun to push the boundaries of model reasoning~\cite{o1report,guo2025deepseekr1,gemini25pro}, aiming to equip models with the ability to synthesize existing knowledge and solve complex, novel problems. For more, readers are referred to this survey~\cite{wang2025multimodal}.

\subsection{Benchmarking Foundation Models} 
Benchmarks serve as an essential tool for evaluating foundation models 
and providing a guiding light for their further development.
Existing benchmarks primarily focus on multi-modal inputs such as text~\cite{hendrycks2021measuring,guo2025rbench,wang2024mmlu,chen2021evaluating,2023opencompass} and multimodality~\cite{yue2024mmmu,yue2024mmmupro,lu2023mathvista,wang2024measuring,lu2023mathvista,gao2024omni}, but their outputs are all textual, overlooking the evaluation of models’ multi-modal output capabilities. 
Besides, some studies~\cite{heusel2017gans,kastryulin2022pytorch,you2024descriptive} have also focused on image generation quality,
but they primarily emphasize aesthetic metrics.

As mentioned in Sec.~\ref{sec_foundation_model}, we believe that the next generation of powerful models are omni-models with strong textual and visual reasoning capabilities.
Thus, in this paper, we present~\method{}, a benchmark for omni reasoning models.
To the best of our knowledge, this is the first attempt to design benchmark to
assess models' multi-modal generation capability in the visual thinking process.

\section{\method{}}

% In this section, we will 
% introduce details of \method{}. It contains the principles for our data collection as well as statistics and examples of proposed benchmark.
% In this section, we will 
% introduce details of \method{},
% including the principles for our data collection as well as statistics and illustrative examples of proposed benchmark.

\subsection{Data Collection of \method{}}

% The core challenge to complete \method{} is 
% How to design suitable questions to assess the
% visual reasoning ability and multi-modal outputs capability in problem-solving process ?

\begin{figure}[t]
\begin{center}
\includegraphics[width=1.0\columnwidth]{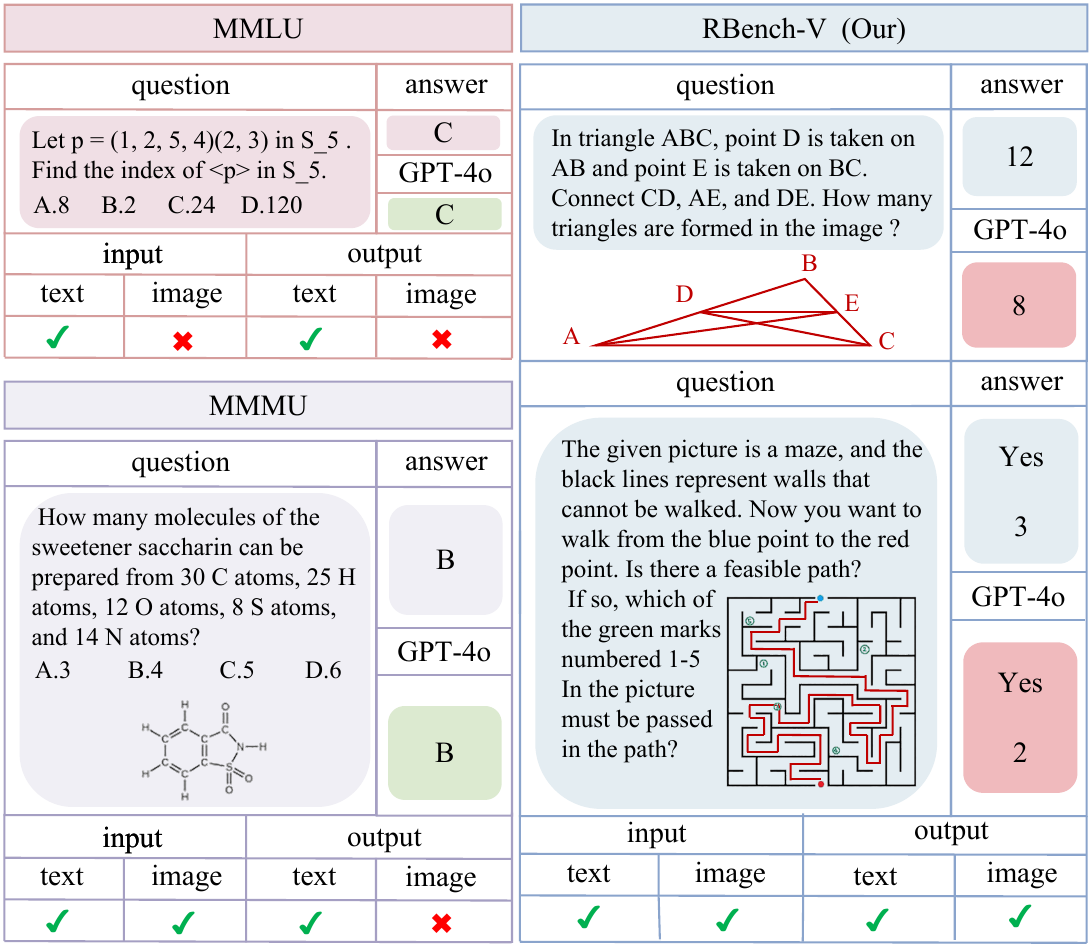}
\caption{A visual comparison with MMLU, MMMU and \method{}. It shows that solving problems in MMLU and MMMU mainly requires understanding multi-modal inputs and generating textual outputs,
whereas solving problems in \method{} demands not only understanding multi-modal inputs but also generating multi-modal outputs.
The \redline{\textbf{red}} lines shown in the figure are \redline{\textbf{not}} part of the original questions and represent the multi-modal reasoning process when solving problems in \method{}, such as drawing geometric shapes or tracing paths through a maze.}
\label{fig_comparsion}
\end{center}
\vskip -0.2in
\end{figure}

The central challenge in developing \method{} lies in designing and curating questions that assess models' ability to generate multi-modal outputs during visual reasoning. 
Clear examples are shown in Fig.~\ref{fig_comparsion}, solving problems in \method{} requires producing outputs beyond text, such as drawing geometric figures (top-right example) or tracing path through a maze (bottom-right example).

To build \method{},
our principle in designing or collecting questions is that their solution should involve creating new visual content, such as creating images or modification of existing images, during the problem-solving process.
We can imagine that numerous real-world scenarios, such as GUI operation and drawing, rely on multi-modal outputs for visual reasoning in daily life. 
In this work, \method{} primarily focuses on math, physics, counting, and games.
To curate high-quality questions in math and physics, we collaborate with domain experts.
For counting and games, we conduct a rigorous rule to create, collect, and filter questions. 
The collection criteria for different domains are detailed as follows.

\paragraph{{\small$\bullet$} \textbf{Math:}} 
For math, we primarily focus on geometric and graph theory problems, including transformation geometry,
planar geometry, solid geometry, \etc.
\textbf{Transformation Geometry:}
The problems in \method{} mainly involve translations, reflections and rotations,  
requiring the model to draw the resulting figures 
after applying these transformations.
in order to arrive at the correct answer.
\textbf{Planar Geometry:} These problems evaluate whether the model can construct appropriate auxiliary lines to aid in reasoning.
\textbf{Solid Geometry:} The solid geometry tasks in \method{} assess  models to assemble 
3D shapes from 2D components according to specific rules and to draw the resulting solid before answering the question.
\textbf{Graph Theory:} \method{} requires models to first complete the graph based on given constraints, mark the leaf nodes, and then reason to arrive at the correct answer.

\paragraph{{\small$\bullet$} \textbf{Physics:}}
We primarily focus on optics, mechanics, electromagnetism, and thermodynamics. 
Not all above problems meet our criteria and we specifically select those that require visual reasoning. \textbf{Optics:} Tasks emphasize geometric optics, requiring models to trace light trajectories involving reflection, refraction, and diffraction. Precise visualization of light paths is essential for deriving optical principles.
\textbf{Mechanics:} This category includes statics, kinematics, and dynamics, involving complex physical constraints. Models must interpret and construct geometric relationships using free-body diagrams and motion trajectories to analyze force interactions, motion paths, and equilibrium conditions. 
\textbf{Electromagnetism:} This area comprises two subcategories. Circuit analysis tasks require models to identify current paths and simplify circuit diagrams in complex scenarios. Dynamic problems combine electromagnetic fields with mechanics, necessitating the visualization of electric and magnetic field lines to analyze particle motion. 
\textbf{Thermodynamics:} Tasks primarily involve fluid force analysis, where models must visually represent dynamic changes in liquid surfaces and force distributions to solve problems related to surface tension, hydrostatic pressure, and buoyancy.

% \paragraph{{\small$\bullet$} \textbf{Chemistry:}}
% \TODO{We primarily focus on several types of games that require visual reasoning and multi-modal outputs: 
% \textbf{Connect-the-dots}: Models need to connect a sequence of dots to reveal an image and identify what object in the image.
% \textbf{Mazes:} Models should trace the correct path through the maze; \method{} formulates questions based on the necessary trajectory.
% \textbf{Ball-and-brick, Dart-and-balloon, and Gold Miner:} These require models to accurately draw the trajectory of a ball, dart, or hook.
% \textbf{Puzzles:} The task involves moving different pieces to complete the full puzzle.
% \textbf{Zuma:} Models need to depict the ball’s movement trajectory and reason about the resulting chain reactions.
% }

\paragraph{{\small$\bullet$} \textbf{Counting:}}
Counting problems typically require no image modification. For instance, identifying two people in a clearly depicted photo needs no interaction.
The counting problems in \method{} differ from the simple example above and can be broadly categorized into the following three types: \textbf{Firstly,} problems require drawing geometric shapes based on descriptions or connecting lines within the image before answering questions such as how many triangles are present (an example is shown in the top-right corner of Fig.~\ref{fig_comparsion}).
\textbf{Secondly,} questions involve images with lots of targets and chaotic backgrounds. Solving such problems requires models to carefully check the images, mark the targets it has counted, and then reason to answer the total. (examples are provided in the supplementary materials.)
\textbf{Thirdly,}  problems demand an understanding of spatial relationships and imagination. Models need to mentally manipulate or move 3D objects and visualize the resulting state after movement, in order to complete the counting task.

% \begin{table}[!t]
% \caption{Statistics on \method{}.}
% \label{tab_stats}
% % \vskip 0.15in
% \begin{center}
% \renewcommand{\tabcolsep}{1.0mm}
% % \begin{small}
% % \begin{sc}
% \begin{tabular}{lc}
% \toprule
% Statistics & Number \\
% % \midrule
% %  & \textbf{Closed-source models}  &  \\
% \midrule
% Total Questions &  1XXX \\
% \midrule
% Math Questions &  1XXX \\
% Physics Questions &  1XXX \\
% Counting Questions &  1XXX \\
% Games Questions &  1XXX \\
% \midrule
% MC Questions  &  1XXX \\
% Open Questions  &  1XXX \\
% \midrule
% Text-only Questions  &  1XXX \\
% multi-modal Questions  &  1XXX \\
% \bottomrule

% \end{tabular}
% % \end{sc}
% % \end{small}
% \end{center}
% \vskip -0.1in
% \end{table}

\paragraph{{\small$\bullet$} \textbf{Games:}} We primarily focus on several types of games that require multi-modal outputs in the visual reasoning process: 
\textbf{Connect-the-dots}: Models need to connect a sequence of dots to reveal an image and identify what object in the image.
\textbf{Mazes:} Models should trace the correct path through the maze and answer questions based on the trajectory.
\textbf{Dart-and-balloon, and Gold Miner:} These require models to accurately draw the trajectory of darts and hooks, and determine 
their intersection with the target objects.
\textbf{Puzzles:} The task involves moving different pieces to complete the full puzzle.
\textbf{Ball-and-brick:} It requires drawing the trajectory of the ball, which may collide and bounce against the wall multiple times.
% \textbf{Darts and balloons:} Models need to simulate the darts‘ flight trajectory to determine which balloons are popped.

\subsection{Statistics of \method{}}

\begin{minipage}[t]{0.6\textwidth}
We conduct statistical analysis on \method{} with the results presented in Tab.~\ref{tab_stats}. 
It presents that \method{} includes 803 questions across four areas, with 176 math questions, 157 physics questions, 195 counting problems, and 275 game-related questions,
comprising 356 multiple-choice questions (We categorize questions with clearly limited answer choices as multiple-choice questions, such as the maze problem in the bottom right of Fig.~\ref{fig_comparsion}.)
 and 447 open-ended questions.
It is worth noting that since we primarily focus on multi-modal outputs rather than inputs, so \method{} includes both text-only and multi-modal input questions. categorized as 40 and 763, respectively.
As an early exploration into visual reasoning and  multi-modal outputs, this paper focuses on text and image modalities, aiming to offer insights for 
foundation models.
As for more modalities such as video and audio outputs, we expect more work to be done in the future.
\end{minipage}%
\hfill
\begin{minipage}[t]{0.38\textwidth}
\captionsetup{type=table}
\captionof{table}{Statistics on \method{}. MC denotes multiple-choice.}
\label{tab_stats}
\centering
\begin{tabular}{@{}lc@{}}
\toprule
\textbf{Statistics} & \textbf{Number} \\
\midrule
Total Questions &  803 \\
\midrule
Math Questions &  176 \\
Physics Questions &  157 \\
Counting Questions &  195 \\
Games Questions &  275 \\
\midrule
MC Questions  &  356 \\
Open Questions  &  447 \\
\midrule
Text-only Questions  &  40 \\
Multi-modal Questions  &  763 \\
\bottomrule
\end{tabular}
\end{minipage}

\section{Experiments}

After developing \method{}, we comprehensively evaluate many open- and closed-source MLLMs 
(\eg Qwen2.5VL~\cite{bai2025qwen2}, Claude3.7~\cite{claude35}, LLaVA-OneVision~\cite{li2024llava}, \etc)
, omni-models (\eg GPT-4o~\cite{gpt4oreport}, Gemini2.5pro~\cite{team2024gemini}, Qwen2.5-Omni~\cite{xu2025qwen2}, \etc) , thinking models (\eg OpenAI o3~\cite{gpto3o4} and Doubao-1.5-thinking-pro-m~\cite{doubaothinking18})
and humam experts on~\method{}. We list all the tested models in Tab.~\ref{tab_main_result}.

By default, all evaluations are conducted in a zero-shot setting.
Besides, since \method{} includes both 
multiple-choice and open-ended questions, 
we adopt a unified LLM-as-a-Judge framework, 
with the judge model being GPT-4o. 
We report Top-1 accuracy (\%) as our default evaluation metric.

\begin{table}
    \caption{Models and experts scores of multi-modal output requirements across different benchmarks.}
    \centering
    \begin{subtable}[t]{0.495\textwidth}
        \centering
        \renewcommand{\tabcolsep}{1.2mm}
        \caption{\method{} vs. MMLU on text-only questions.}
        \begin{tabular}{lccc}
            \hline
            & \method{} win & MMLU win & Tie \\
            \hline
            Model  &  92.4  &  3.8  &  3.8 \\
            Human  &  86.4  &  4.5  &  9.1 \\
            \hline
        \end{tabular}
    \end{subtable}
    \hfill
    \begin{subtable}[t]{0.495\textwidth}
        \centering
        \renewcommand{\tabcolsep}{1.2mm}
        \caption{\method{} vs. MMMU on multi-modal questions.}
        \begin{tabular}{lccc}
            \hline
            & \method{} win & MMMU win & Tie \\
            \hline
            Model  &  94.1  &  5.9  &  0.0  \\
            Human  &  83.4  &  6.6  &  10.0 \\
            \hline
        \end{tabular}
    \end{subtable}
\label{tab_user_study}
\end{table}

\subsection{Comparison with other benchmarks on multi-modal output capability requirements}
Here, we compare \method{} with other benchmarks (MMLU, MMMU)
in terms of multi-modal output evaluation. 
As we know, it is challenging to find a quantitative method to assess the property for multi-modal outputs in existing benchmarks.
Therefore, we construct expert scores and model scores to measure their multi-modal reasoning property.
% we conduct user study, inviting senior undergraduate students to judge whether solving questions from different benchmarks require multi-modal outputs.
% Besides, 
% we also ask foundation models~\cite{} to answer above questions. 
Specifically, we randomly sample 30 examples from each of \method{}, MMLU, and MMMU, and instruct either human experts or the models (o3 and Doubao-1.5-thinking-m) to compare whether requires drawing during the thinking process.

We summarize the win rates in Table~\ref{tab_user_study}.
Results from both human experts and models consistently show that \method{} imposes a significantly higher requirement for multi-modal outputs during the reasoning process compared to MMLU and MMMU.
This highlights that \method{} is specifically designed to assess multi-modal output capabilities and visual reasoning skills.

% \begin{figure}[t]
% \begin{center}
% \begin{overpic}[width=1.0\linewidth]{./images/teaser_v2.pdf}
%   \put(17, 0){(a) Error Statistics by Type}
%   \put(65, 0){(b) Error Statistics by Difficulty}
% \end{overpic}\hspace{1pt}
% % \includegraphics[width=1.0\columnwidth]{./images/teaser_v2.pdf}
% \caption{Placeholder XXXXXX.}
% \label{fig_error}
% \end{center}
% \vskip -0.2in
% \end{figure}

\subsection{Evaluating visual reasoning and multi-modal outputs of different models}
We assess various open- and closed-source models, 
along with human experts, on \method{}. 
The specific models are listed in Tab.~\ref{tab_main_result}.
For open-source models, we use vLLM~\cite{kwon2023efficient} and VLMEvalKit~\cite{duan2024vlmevalkit} for deployment, setting the temperature as 0 
while leaving all other parameters at their default values.
The above experiments are conducted on 8 x NVIDIA H20 GPUs.
For closed-source models, 
we follow the official API usage guidelines provided for each model. 
If the official API allows setting the temperature parameter, we set it to 0; all other parameters are kept as recommended by the official documentation.
For the human expert score, we invite senior undergraduate students to serve as human experts. For physics and math, we recruit some senior undergraduates for each major, assigning them different sets of questions.
For games and counting tasks, we similarly invite some senior undergraduates, without restricting their academic backgrounds.
The experimental results are summarized in Tab.~\ref{tab_main_result}, and a comprehensive analysis is presented in Sec.~\ref{finding}.

% \subsection{Case analysis}

% \subsubsection{Error analysis}

% \TODO{We randomly select 100 incorrect examples from GPT-4o for analysis and categorized the types of errors. The results are presented in Tab.~\ref{XXX}.
% We randomly select 100 incorrect examples from GPT-4o for analysis and categorized the types of errors. The results are presented in Tab.~\ref{XXX}.}

\begin{table}[t]
\caption{Performance (\%) of different models and human experts on \method{}. $^\dag$ means long chain thinking model.  $^*$ represents the omni-model. The highest scores are highlighted in \textbf{\red{red}}, and the second-highest scores are highlighted in \textbf{\blue{blue}}.}
\label{tab_main_result}
% \vskip 0.15in
\begin{center}
\renewcommand{\tabcolsep}{0.5mm}
% \begin{small}
% \begin{sc}
\begin{tabular}{l|c|c|cccc}
\toprule
Name  & Overall & w/o Math & Math & Physics & Counting & Games \\
% \midrule
%  & \textbf{Closed-source models}  &  \\
\midrule
\rowcolor{gray!20}
\multicolumn{7}{c}{Open-source models} \\
% \midrule
Qwen2.5-Omni-7B$^{*}$
~\cite{xu2025qwen2} &  7.7 & 4.5 & 11.4 & 1.9 & 2.1  & 7.7   \\
InternVL-3-14B~\cite{zhu2025internvl3} &  8.0 & 7.0 & 11.4 & 1.3 & 5.1 & 11.6  \\
InternVL-3-8B~\cite{zhu2025internvl3} &  8.2 & 6.0 & 15.9 & 1.9 & 5.6 & 8.7   \\
Qwen2.5VL-7B~\cite{bai2025qwen2} &  8.3 & 7.0 & 13.1 & 2.5  &  3.6 &  12.0 \\
LLaVA-OneVision-7B~\cite{li2024llava} &  8.5  & 6.8 & 14.2 & 2.5 & 4.6 & 10.9 \\
DeepSeek-VL2~\cite{wu2024deepseek} &  9.0 & 7.0 & 15.9 & 0.6 & 5.6 & 11.6  \\
LLaVA-OneVision-72B~\cite{li2024llava} &  9.0  & 8.9 & 9.1 & 4.5 & 4.6 & 14.5 \\
MiniCPM-2.6-V~\cite{yao2024minicpm} & 9.1 & 7.2 & 15.9  & 1.3  & 6.2  &  11.3   \\
Llama4-Scout (109B MoE)~\cite{llama4} &  9.5  & 6.9 & 18.8  &  3.2 &  4.1  & 10.9 \\
MiniCPM-2.6-o$^{*}$~\cite{yao2024minicpm} &  9.7 &  7.5  & 17.6 & 1.3 & 3.6 & 13.8 \\
Qwen2.5VL-32B~\cite{bai2025qwen2} &  10.0  & 6.4 & 22.7  & 2.5  & 4.1 & 10.2 \\
InternVL-3-38B~\cite{zhu2025internvl3} &  10.0 & 7.2 & 20.5 & 0.6 & 5.1 & 12.4  \\
Qwen2.5VL-72B~\cite{bai2025qwen2} &  10.6  & 9.2 & 15.3  & 3.8  & 6.2 & 14.5 \\
\midrule
\rowcolor{gray!20}
\multicolumn{7}{c}{Closed-source models} \\
% \midrule
QVQ-Max~\cite{qvqmax} &  11.0  & 8.1 & 21.0 &  5.7 & 6.2 & 10.9 \\
Claude-3.7-sonnet~\cite{claude37} &  11.5 & 9.1 & 19.9 & 3.8 & 8.7 & 12.4  \\
OpenAI GPT-4.5~\cite{gpt45} &12.6  & 11.0  & 18.2 & 2.5 & 11.8 & 15.3 \\
Step-R1-V-Mini$^\dag$~\cite{stepr1v} &  13.2  & 8.8  & 29.0 & 6.4 & 10.3 & 9.1 \\
OpenAI GPT-4.1~\cite{gpt41report} & 13.6  & 11.7 & 20.5 & 5.7 & 11.3 & 15.3 \\
OpenAI GPT-4o-20250327$^{*}$~\cite{gpt4oreport} &  14.1  & 11.2  & 24.4 & 3.2 & 13.3  & 14.2 \\
Doubao-1.5-vision-pro~\cite{doubao15} &  15.6  & 11.5  & 30.1 & 8.9 & 12.8 & 12.0 \\
OpenAI o1$^\dag$~\cite{o1report} &  16.2  & 11.0 &  34.7 & 5.7 &  12.3 & 13.1 \\
Doubao-1.5-thinking-pro-m$^\dag$~\cite{doubaothinking18} &  17.1 & 11.0  & 38.6  &  13.4 & 9.7 & 10.5  \\
Gemini 2.5 pro-preview-0506 $^{*}$ ~\cite{gemini25pro} &  20.2 & 13.9 & 42.6  & 9.6  & 19.0  & 12.7 \\
OpenAI o4-mini$^\dag$~\cite{gpto3o4} &  20.9  & 14.6 & 43.2 & 12.7 & 17.4 & 13.8 \\
OpenAI o3$^\dag$~\cite{gpto3o4} &  \textbf{\blue{25.8}}   &  \textbf{\blue{19.5}} & \textbf{\blue{48.3}} & \textbf{\blue{20.4}}  &   \textbf{\blue{22.1}} & \textbf{\blue{17.1}} \\
\midrule
\rowcolor{gray!20}
\multicolumn{7}{c}{Human Experts} \\
% \midrule
% \rowcolor{green!15}
Human Experts Score & \textbf{\red{82.3}} &  \textbf{\red{81.7}} & \textbf{\red{84.7}} & \textbf{\red{69.4}} & \textbf{\red{81.0}}  &  \textbf{\red{89.1}}   \\
\bottomrule
\end{tabular}
% \end{sc}
% \end{small}
\end{center}
\vskip -0.1in
\end{table}

\subsection{Visualization}

\begin{figure}[t]
\begin{center}
\includegraphics[width=1.0\columnwidth]{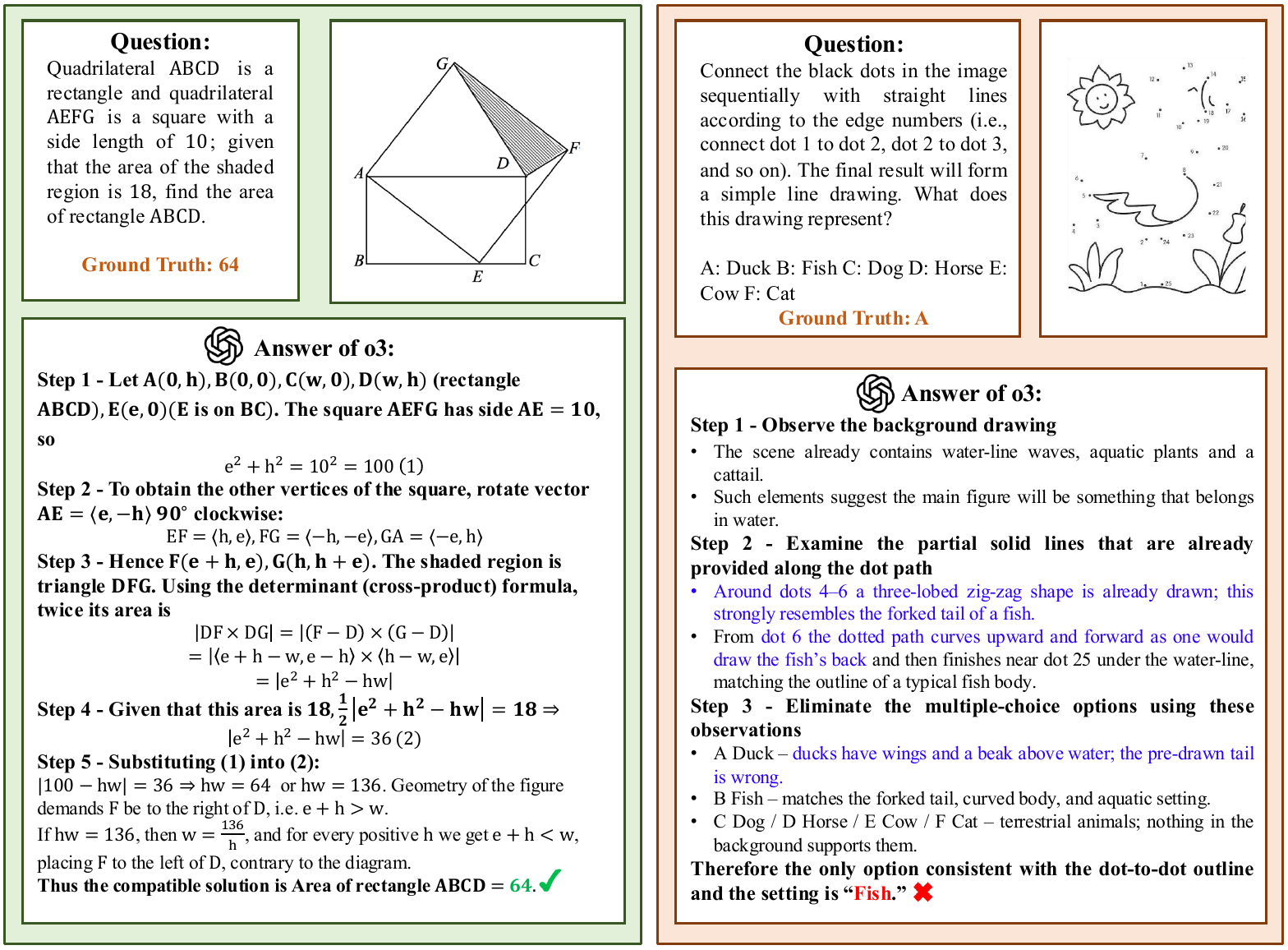}
\caption{Examples of o3's responses to math and game questions in \method{}.
Left: o3 correctly answers a math question in \method{} by transforming the geometry problem into an algebraic one using a coordinate system, whereas humans typically solve it using geometric methods.
Right: o3 fails to answer a game question correctly. The \blue{blue} highlights indicate the cause of the error and the key issue is that the model fails to follow the instructions to draw the required connections.}
\label{fig_case_study}
\end{center}
% \vskip -0.2in
\end{figure}

Here, we visualize a correct example and an incorrect example: the correct case is from planar geometry in math,
while the incorrect case is from games. 
The answer comes from the representative model, o3, the results are shown in Fig.~\ref{fig_case_study}.

From the results, we observe that although o3 arrives at the correct answer for the planar geometry question, its solution is based on an algebraic approach by establishing a coordinate system, rather than using a typical geometric method commonly adopted by humans. 
This suggests that the model tends to favor algebraic solutions with text-only reasoning thought over multi-modal geometric reasoning path when both approaches are available.
It indicates that improvements in mathematical performance do not necessarily reflect genuine advancements in multimodal reasoning ability, but rather suggest that models have learned certain “multi-modal reasoning shortcuts”.
Experts in mathematics have validated this hypothesis, pointing out that most geometry problems can be solved using algebraic methods.
In contrast, counting, and games do not exhibit such “multimodal reasoning shortcuts“.
Therefore, we also report the performance under the “w/o math” setting in Table~\ref{tab_main_result}, which may serve as a better indicator of a model’s true multimodal reasoning capability.

In the second question, derived from a connect-the-dots task in the games category, o3 fails to generate a correct answer.
Analysis reveals that the errors here mainly stem from o3 merely attempting to describe the points in the diagram, rather than actually connecting them as required by the question.
Due to space limitations, we are unable to present more examples, but our analysis shows that the majority of model failures are caused by this limitation.

\subsection{Observation and findings}
\label{finding}

\textit{If models, such as the InternVL or Qwen-VL series, lack multi-modal CoT, merely increasing their model sizes will not effectively resolve the challenge of visual reasoning.}
As shown in Tab.~\ref{tab_main_result}, increasing the parameter size of the Qwen2.5VL model from 7B to 72B does not result in a clear performance improvement on \method{}. 
A similar phenomenon is also observed in the InternVL and LLaVA-OneVision series.
It suggests that the scaling law may be insufficient to address the challenges of multi-modal output in visual reasoning.
Furthermore, we question whether foundation models trained primarily via next-token prediction are inherently limited in their ability to handle such tasks. 
While this training paradigm is well-suited for text generation, it may be fundamentally inadequate for detailed and precise multi-modal generation and understanding such as precisely tracing curve trajectory in mazes.

\textit{Omni-models and long text-only CoT approaches also do not show significant improvements on this task.
}
As shown in Table~\ref{tab_main_result}, a comparison between Qwen2.5VL-7B and Qwen2.5VL-Omni-7B, as well as between MiniCPM-V-2.6 and MiniCPM-o-2.6, indicates that simply incorporating images during output decoding,
as done in omni-models, 
fails to effectively address the multimodal reasoning challenge.
In addition, typical long text-only thinking models also show only marginal gains on \method{}, as evidenced by the comparison between Double1.5-vision-pro and Doubao1.5-thinking-pro-m.
Combining our analysis on scaling laws, omni-models, and long text-only CoT approaches,
indicating that for current foundation models, novel techniques such as M-CoT and agents can be required to effectively solve visual reasoning tasks involving precise multimodal outputs.

% \textit{Foundation models still fall well short of human expert performance in generating multi-modal outputs during visual reasoning.}
% As shown in Tab.~\ref{tab_main_result}, 
% even the best-performing model to date, o3, only achieves an overall accuracy of 25.8\%, which still lags significantly behind the human expert score of 82.3\%. 
% This phenomenon is clearly illustrated by the bar chart in Fig.~\ref{fig_comparsion_company},
% highlighting that there is still considerable room for improvement in visual reasoning tasks requiring multimodal outputs.

\textit{Foundation models still fall well short of human expert performance in generating multi-modal outputs during visual reasoning.}
As shown in Table~\ref{tab_main_result}, even the best-performing model to date, o3, achieves only an overall accuracy of 25.8\%, which remains significantly behind the human expert score of 82.3\%.
This substantial performance gap underscores the limitations of current foundation models in handling tasks that demand precise multi-modal outputs in the visual reasoning process.
This phenomenon is clearly illustrated by the bar chart in Fig.~\ref{fig_comparsion_company}. The results emphasize that, despite recent progress, there is still considerable room for advancement in multimodal reasoning.

\textit{The methods used by human experts and models to solve problems are not consistent.}
As shown in the Tab.~\ref{tab_main_result}, various models tend to perform best on the mathematics subject. We analyze and present representative examples from mathematics in Fig.~\ref{fig_case_study}, revealing that models often convert geometric problems into algebraic ones by constructing coordinate systems.
This approach differs significantly from that of human experts, who typically solve such problems using geometric reasoning.
It suggests that the intelligence of current models differs from that of humans.
Therefore, to avoid such “multi-modal reasoning shortcuts,” we also report the accuracy after removing math-related questions in Tab~\ref{tab_main_result}.
The results show that excluding math further amplifies the performance gap between models and human experts.

% , and that new technical solutions may be necessary. 
% We also discussed from a mechanistic explanation in the previous finding.
% A similar trend is also observed in the InternVL3 model series.

\textit{OpenAI o3 has made substantial progress in visual reasoning with multi-modal output.}
The release of o3 attracted widespread attention, largely due to its impressive ability to handle complex visual reasoning tasks—a capability that has been challenging for previous models.
We observed the same phenomenon in our proposed \method{}, where o3 significantly outperformed all other models in tasks that require accurate and coherent multimodal outputs.
This performance lead suggests that o3 has undergone deliberate and effective enhancements specifically aimed at improving its visual reasoning and output alignment capabilities.
Notably, the results also validate the design of \method{} itself.
It demonstrates \method{} can serve as a reliable benchmark for evaluating progress and tracking how models are evolving toward human-level multimodal reasoning.
% It also demonstrates that \method{} serves as an effective evaluation tool, both for tracking the progress of models aiming to catch up with o3 and for assessing future improvements to o3 itself.

% The release of OpenAI's o3 model caused a significant public stir due to its remarkable visual reasoning capabilities. We observed the same phenomenon in our proposed \method{}, 
% where o3 substantially outperformed all other models. We believe this indicates that o3 has undergone targeted improvements in this area.

\textit{Open-source models still lag far behind closed-source models.}
Although open-source models such as Qwen2.5VL~\cite{bai2025qwen2} and LLaMA 4~\cite{llama4} are making continuous progress, there remains a noticeable gap (10.6\% vs. 25.8\%) between open-source and closed-source models in visual reasoning tasks that require multi-modal outputs. 
In addition, we find that the performance of current open-source models is quite similar, with low accuracy rates mostly ranging between 8\% and 10\%. 
It suggests that current open-source models exhibit only minimal capability in multi-modal reasoning.
We hope the community will develop new techniques based on open-source models to enhance their multi-modal output capabilities and ultimately close the gap with closed-source models on \method{}.

\section{Conclusion}

In this work, we carefully hand-pick 803 questions across 4 topics and propose \method{}, a benchmark specifically designed to evaluate models' multimodal output capabilities in the visual reasoning process. 
It systematically assesses the current performance of models, highlights the progress made by the o3 model in this domain, 
and reveals the significant gap between current intelligent models and human experts. 
Besides, according to our observation, 
the current technologies such as scaling law, long text-only CoT and joint text-visual decoding, fail to effectively address the challenges posed by \method{}.
% Furthermore, it provides guidance for the future development of foundation models. 

Looking ahead, we hope that foundation models will evolve towards the direction of omni reasoning models and achieve better performance on \method{}.
In addition, we will also try to improve the foundation models from the perspective of M-CoT and the agent.

\bibliographystyle{plainnat} % NeurIPS 推荐使用 plainnat 风格
\bibliography{references} % 文件名，不需要加 .bib 后缀

\begin{thebibliography}{52}
\providecommand{\natexlab}[1]{#1}
\providecommand{\url}[1]{\texttt{#1}}
\expandafter\ifx\csname urlstyle\endcsname\relax
  \providecommand{\doi}[1]{doi: #1}\else
  \providecommand{\doi}{doi: \begingroup \urlstyle{rm}\Url}\fi

\bibitem[Abdin et~al.(2024)Abdin, Aneja, Awadalla, Awadallah, Awan, Bach, Bahree, Bakhtiari, Bao, Behl, et~al.]{abdin2024phi}
Marah Abdin, Jyoti Aneja, Hany Awadalla, Ahmed Awadallah, Ammar~Ahmad Awan, Nguyen Bach, Amit Bahree, Arash Bakhtiari, Jianmin Bao, Harkirat Behl, et~al.
\newblock Phi-3 technical report: A highly capable language model locally on your phone.
\newblock \emph{arXiv preprint arXiv:2404.14219}, 2024.

\bibitem[Anthropic(2024)]{claude35}
Anthropic.
\newblock Introducing claude 3.5 sonnet.
\newblock \emph{https://www.anthropic.com/news/claude-3-5-sonnet}, 2024.

\bibitem[Anthropic(2025)]{claude37}
Anthropic.
\newblock Claude 3.7 sonnet.
\newblock \emph{https://www.anthropic.com/claude/sonnet}, 2025.

\bibitem[Bai et~al.(2023)Bai, Bai, Chu, Cui, Dang, Deng, Fan, Ge, Han, Huang, et~al.]{bai2023qwen}
Jinze Bai, Shuai Bai, Yunfei Chu, Zeyu Cui, Kai Dang, Xiaodong Deng, Yang Fan, Wenbin Ge, Yu~Han, Fei Huang, et~al.
\newblock Qwen technical report.
\newblock \emph{arXiv preprint arXiv:2309.16609}, 2023.

\bibitem[Bai et~al.(2025)Bai, Chen, Liu, Wang, Ge, Song, Dang, Wang, Wang, Tang, et~al.]{bai2025qwen2}
Shuai Bai, Keqin Chen, Xuejing Liu, Jialin Wang, Wenbin Ge, Sibo Song, Kai Dang, Peng Wang, Shijie Wang, Jun Tang, et~al.
\newblock Qwen2.5-vl technical report.
\newblock \emph{arXiv preprint arXiv:2502.13923}, 2025.

\bibitem[Chen et~al.(2021)Chen, Tworek, Jun, Yuan, Pinto, Kaplan, Edwards, Burda, Joseph, Brockman, et~al.]{chen2021evaluating}
Mark Chen, Jerry Tworek, Heewoo Jun, Qiming Yuan, Henrique Ponde De~Oliveira Pinto, Jared Kaplan, Harri Edwards, Yuri Burda, Nicholas Joseph, Greg Brockman, et~al.
\newblock Evaluating large language models trained on code.
\newblock \emph{arXiv preprint arXiv:2107.03374}, 2021.

\bibitem[Contributors(2023)]{2023opencompass}
OpenCompass Contributors.
\newblock Opencompass: A universal evaluation platform for foundation models.
\newblock \url{https://github.com/open-compass/opencompass}, 2023.

\bibitem[DeepSeek(2025)]{guo2025deepseekr1}
DeepSeek.
\newblock Deepseek-r1: Incentivizing reasoning capability in llms via reinforcement learning.
\newblock \emph{arXiv preprint arXiv:2501.12948}, 2025.

\bibitem[Duan et~al.(2024)Duan, Yang, Qiao, Fang, Chen, Liu, Dong, Zang, Zhang, Wang, et~al.]{duan2024vlmevalkit}
Haodong Duan, Junming Yang, Yuxuan Qiao, Xinyu Fang, Lin Chen, Yuan Liu, Xiaoyi Dong, Yuhang Zang, Pan Zhang, Jiaqi Wang, et~al.
\newblock Vlmevalkit: An open-source toolkit for evaluating large multi-modality models.
\newblock In \emph{Proceedings of the 32nd ACM international conference on multimedia}, pages 11198--11201, 2024.

\bibitem[Edwards(2012)]{edwards2012drawing}
Betty Edwards.
\newblock \emph{Drawing on the right side of the brain: The definitive}.
\newblock Penguin, 2012.

\bibitem[Fan et~al.(2023)Fan, Bainbridge, Chamberlain, and Wammes]{fan2023drawing}
Judith~E Fan, Wilma~A Bainbridge, Rebecca Chamberlain, and Jeffrey~D Wammes.
\newblock Drawing as a versatile cognitive tool.
\newblock \emph{Nature Reviews Psychology}, 2\penalty0 (9):\penalty0 556--568, 2023.

\bibitem[Gao et~al.(2024)Gao, Song, Yang, Cai, Miao, Dong, Li, Ma, Chen, Xu, et~al.]{gao2024omni}
Bofei Gao, Feifan Song, Zhe Yang, Zefan Cai, Yibo Miao, Qingxiu Dong, Lei Li, Chenghao Ma, Liang Chen, Runxin Xu, et~al.
\newblock Omni-math: A universal olympiad level mathematic benchmark for large language models.
\newblock \emph{arXiv preprint arXiv:2410.07985}, 2024.

\bibitem[Goldschmidt(1991)]{goldschmidt1991dialectics}
Gabriela Goldschmidt.
\newblock The dialectics of sketching.
\newblock \emph{Creativity research journal}, 4\penalty0 (2):\penalty0 123--143, 1991.

\bibitem[Google(2025)]{gemini25pro}
Google.
\newblock Gemini 2.5: Our most intelligent ai model.
\newblock \emph{https://blog.google/technology/google-deepmind/gemini-model-thinking-updates-march-2025/\#gemini-2-5-thinking}, 2025.

\bibitem[Google et~al.(2023)Google, Anil, Borgeaud, Alayrac, Yu, Soricut, Schalkwyk, Dai, Hauth, Millican, et~al.]{team2023gemini}
Google, Rohan Anil, Sebastian Borgeaud, Jean-Baptiste Alayrac, Jiahui Yu, Radu Soricut, Johan Schalkwyk, Andrew~M Dai, Anja Hauth, Katie Millican, et~al.
\newblock Gemini: a family of highly capable multimodal models.
\newblock \emph{arXiv preprint arXiv:2312.11805}, 2023.

\bibitem[Google et~al.(2024)Google, Georgiev, Lei, Burnell, Bai, Gulati, Tanzer, Vincent, Pan, Wang, et~al.]{team2024gemini}
Google, Petko Georgiev, Ving~Ian Lei, Ryan Burnell, Libin Bai, Anmol Gulati, Garrett Tanzer, Damien Vincent, Zhufeng Pan, Shibo Wang, et~al.
\newblock Gemini 1.5: Unlocking multimodal understanding across millions of tokens of context.
\newblock \emph{arXiv preprint arXiv:2403.05530}, 2024.

\bibitem[Guo et~al.(2025)Guo, Xu, Zhang, Song, Peng, Deng, Dong, Nakayama, Geng, Wang, Ni, Yang, Rao, Peng, Hu, Wetzstein, and min Hu]{guo2025rbench}
Meng-Hao Guo, Jiajun Xu, Yi~Zhang, Jiaxi Song, Haoyang Peng, Yi-Xuan Deng, Xinzhi Dong, Kiyohiro Nakayama, Zhengyang Geng, Chen Wang, Bolin Ni, Guo-Wei Yang, Yongming Rao, Houwen Peng, Han Hu, Gordon Wetzstein, and Shi min Hu.
\newblock R-bench: Graduate-level multi-disciplinary benchmarks for llm \& mllm complex reasoning evaluation, 2025.
\newblock URL \url{https://arxiv.org/abs/2505.02018}.

\bibitem[Hendrycks et~al.(2021)Hendrycks, Burns, Basart, Zou, Mazeika, Song, and Steinhardt]{hendrycks2021measuring}
Dan Hendrycks, Collin Burns, Steven Basart, Andy Zou, Mantas Mazeika, Dawn Song, and Jacob Steinhardt.
\newblock Measuring massive multitask language understanding.
\newblock In \emph{International Conference on Learning Representations}, 2021.
\newblock URL \url{https://openreview.net/forum?id=d7KBjmI3GmQ}.

\bibitem[Heusel et~al.(2017)Heusel, Ramsauer, Unterthiner, Nessler, and Hochreiter]{heusel2017gans}
Martin Heusel, Hubert Ramsauer, Thomas Unterthiner, Bernhard Nessler, and Sepp Hochreiter.
\newblock Gans trained by a two time-scale update rule converge to a local nash equilibrium.
\newblock \emph{Advances in neural information processing systems}, 30, 2017.

\bibitem[Jiang et~al.(2023)Jiang, Sablayrolles, Mensch, Bamford, Chaplot, Casas, Bressand, Lengyel, Lample, Saulnier, et~al.]{jiang2023mistral}
Albert~Q Jiang, Alexandre Sablayrolles, Arthur Mensch, Chris Bamford, Devendra~Singh Chaplot, Diego de~las Casas, Florian Bressand, Gianna Lengyel, Guillaume Lample, Lucile Saulnier, et~al.
\newblock Mistral 7b.
\newblock \emph{arXiv preprint arXiv:2310.06825}, 2023.

\bibitem[Kastryulin et~al.(2022)Kastryulin, Zakirov, Prokopenko, and Dylov]{kastryulin2022pytorch}
Sergey Kastryulin, Jamil Zakirov, Denis Prokopenko, and Dmitry~V Dylov.
\newblock Pytorch image quality: Metrics for image quality assessment.
\newblock \emph{arXiv preprint arXiv:2208.14818}, 2022.

\bibitem[Kwon et~al.(2023)Kwon, Li, Zhuang, Sheng, Zheng, Yu, Gonzalez, Zhang, and Stoica]{kwon2023efficient}
Woosuk Kwon, Zhuohan Li, Siyuan Zhuang, Ying Sheng, Lianmin Zheng, Cody~Hao Yu, Joseph~E. Gonzalez, Hao Zhang, and Ion Stoica.
\newblock Efficient memory management for large language model serving with pagedattention.
\newblock In \emph{Proceedings of the ACM SIGOPS 29th Symposium on Operating Systems Principles}, 2023.

\bibitem[Li et~al.(2024)Li, Zhang, Guo, Zhang, Li, Zhang, Zhang, Zhang, Li, Liu, et~al.]{li2024llava}
Bo~Li, Yuanhan Zhang, Dong Guo, Renrui Zhang, Feng Li, Hao Zhang, Kaichen Zhang, Peiyuan Zhang, Yanwei Li, Ziwei Liu, et~al.
\newblock Llava-onevision: Easy visual task transfer.
\newblock \emph{arXiv preprint arXiv:2408.03326}, 2024.

\bibitem[Liu et~al.(2024)Liu, Li, Wu, and Lee]{liu2024visual}
Haotian Liu, Chunyuan Li, Qingyang Wu, and Yong~Jae Lee.
\newblock Visual instruction tuning.
\newblock \emph{Advances in neural information processing systems}, 36, 2024.

\bibitem[Lu et~al.(2023)Lu, Bansal, Xia, Liu, Li, Hajishirzi, Cheng, Chang, Galley, and Gao]{lu2023mathvista}
Pan Lu, Hritik Bansal, Tony Xia, Jiacheng Liu, Chunyuan Li, Hannaneh Hajishirzi, Hao Cheng, Kai-Wei Chang, Michel Galley, and Jianfeng Gao.
\newblock Mathvista: Evaluating mathematical reasoning of foundation models in visual contexts.
\newblock \emph{arXiv preprint arXiv:2310.02255}, 2023.

\bibitem[Meta(2025)]{llama4}
Meta.
\newblock The llama 4 herd: The beginning of a new era of natively multimodal ai innovation.
\newblock \emph{https://ai.meta.com/blog/llama-4-multimodal-intelligence/}, 2025.

\bibitem[OpenAI(2022)]{chatgpt}
OpenAI.
\newblock Introducing chatgpt.
\newblock \emph{https://openai.com/index/chatgpt/}, 2022.

\bibitem[OpenAI(2023)]{gpt4vreport}
OpenAI.
\newblock Gpt-4v(ision) system card.
\newblock \emph{https://openai.com/index/gpt-4v-system-card/}, 2023.

\bibitem[OpenAI(2024{\natexlab{a}})]{gpt4oreport}
OpenAI.
\newblock Gpt-4o system card.
\newblock \emph{https://openai.com/index/gpt-4o-system-card/}, 2024{\natexlab{a}}.

\bibitem[OpenAI(2024{\natexlab{b}})]{o1report}
OpenAI.
\newblock Learning to reason with llms.
\newblock \emph{https://openai.com/index/learning-to-reason-with-llms/}, 2024{\natexlab{b}}.

\bibitem[OpenAI(2025{\natexlab{a}})]{gpt41report}
OpenAI.
\newblock Introducing gpt-4.1 in the api.
\newblock \emph{https://openai.com/index/gpt-4-1/}, 2025{\natexlab{a}}.

\bibitem[OpenAI(2025{\natexlab{b}})]{gpt45}
OpenAI.
\newblock Introducing gpt 4.5.
\newblock \emph{https://openai.com/index/introducing-gpt-4-5/}, 2025{\natexlab{b}}.

\bibitem[OpenAI(2025{\natexlab{c}})]{gpto3o4}
OpenAI.
\newblock Openai o3 and o4-mini system card.
\newblock \emph{https://openai.com/index/o3-o4-mini-system-card/}, 2025{\natexlab{c}}.

\bibitem[Pylyshyn(2001)]{pylyshyn2001visual}
Zenon~W Pylyshyn.
\newblock Visual indexes, preconceptual objects, and situated vision.
\newblock \emph{Cognition}, 80\penalty0 (1-2):\penalty0 127--158, 2001.

\bibitem[Qwen(2025)]{qvqmax}
Qwen.
\newblock Qvq-max: Think with evidence.
\newblock \emph{https://qwenlm.github.io/blog/qvq-max-preview/}, 2025.

\bibitem[Seed(2025{\natexlab{a}})]{doubao15}
Seed.
\newblock Doubao-1.5-pro.
\newblock \emph{\url{https://seed.bytedance.com/en/special/doubao\_1\_5\_pro/}}, 2025{\natexlab{a}}.

\bibitem[Seed(2025{\natexlab{b}})]{doubaothinking18}
Seed.
\newblock Seed1.5-thinking: Advancing superb reasoning models with reinforcement learning, 2025{\natexlab{b}}.
\newblock URL \url{https://arxiv.org/abs/2504.13914}.

\bibitem[StepFun(2025)]{stepr1v}
StepFun.
\newblock Step-r1-v-mini.
\newblock \emph{https://www.stepfun.com/chats/new}, 2025.

\bibitem[Touvron et~al.(2023)Touvron, Lavril, Izacard, Martinet, Lachaux, Lacroix, Rozi{\`e}re, Goyal, Hambro, Azhar, et~al.]{touvron2023llama}
Hugo Touvron, Thibaut Lavril, Gautier Izacard, Xavier Martinet, Marie-Anne Lachaux, Timoth{\'e}e Lacroix, Baptiste Rozi{\`e}re, Naman Goyal, Eric Hambro, Faisal Azhar, et~al.
\newblock Llama: Open and efficient foundation language models.
\newblock \emph{arXiv preprint arXiv:2302.13971}, 2023.

\bibitem[Wang et~al.(2024{\natexlab{a}})Wang, Pan, Shi, Lu, Zhan, and Li]{wang2024measuring}
Ke~Wang, Junting Pan, Weikang Shi, Zimu Lu, Mingjie Zhan, and Hongsheng Li.
\newblock Measuring multimodal mathematical reasoning with math-vision dataset.
\newblock \emph{arXiv preprint arXiv:2402.14804}, 2024{\natexlab{a}}.

\bibitem[Wang et~al.(2024{\natexlab{b}})Wang, Zhang, Luo, Sun, Cui, Wang, Zhang, Wang, Li, Yu, et~al.]{wang2024emu3}
Xinlong Wang, Xiaosong Zhang, Zhengxiong Luo, Quan Sun, Yufeng Cui, Jinsheng Wang, Fan Zhang, Yueze Wang, Zhen Li, Qiying Yu, et~al.
\newblock Emu3: Next-token prediction is all you need.
\newblock \emph{arXiv preprint arXiv:2409.18869}, 2024{\natexlab{b}}.

\bibitem[Wang et~al.(2025)Wang, Wu, Zhang, Yan, Liu, Luo, and Fei]{wang2025multimodal}
Yaoting Wang, Shengqiong Wu, Yuecheng Zhang, Shuicheng Yan, Ziwei Liu, Jiebo Luo, and Hao Fei.
\newblock Multimodal chain-of-thought reasoning: A comprehensive survey.
\newblock \emph{arXiv preprint arXiv:2503.12605}, 2025.

\bibitem[Wang et~al.(2024{\natexlab{c}})Wang, Ma, Zhang, Ni, Chandra, Guo, Ren, Arulraj, He, Jiang, et~al.]{wang2024mmlu}
Yubo Wang, Xueguang Ma, Ge~Zhang, Yuansheng Ni, Abhranil Chandra, Shiguang Guo, Weiming Ren, Aaran Arulraj, Xuan He, Ziyan Jiang, et~al.
\newblock Mmlu-pro: A more robust and challenging multi-task language understanding benchmark.
\newblock \emph{arXiv preprint arXiv:2406.01574}, 2024{\natexlab{c}}.

\bibitem[Wu et~al.(2024)Wu, Chen, Pan, Liu, Liu, Dai, Gao, Ma, Wu, Wang, et~al.]{wu2024deepseek}
Zhiyu Wu, Xiaokang Chen, Zizheng Pan, Xingchao Liu, Wen Liu, Damai Dai, Huazuo Gao, Yiyang Ma, Chengyue Wu, Bingxuan Wang, et~al.
\newblock Deepseek-vl2: Mixture-of-experts vision-language models for advanced multimodal understanding.
\newblock \emph{arXiv preprint arXiv:2412.10302}, 2024.

\bibitem[Xu et~al.(2025)Xu, Guo, He, Hu, He, Bai, Chen, Wang, Fan, Dang, et~al.]{xu2025qwen2}
Jin Xu, Zhifang Guo, Jinzheng He, Hangrui Hu, Ting He, Shuai Bai, Keqin Chen, Jialin Wang, Yang Fan, Kai Dang, et~al.
\newblock Qwen2. 5-omni technical report.
\newblock \emph{arXiv preprint arXiv:2503.20215}, 2025.

\bibitem[Yao et~al.(2024)Yao, Yu, Zhang, Wang, Cui, Zhu, Cai, Li, Zhao, He, et~al.]{yao2024minicpm}
Yuan Yao, Tianyu Yu, Ao~Zhang, Chongyi Wang, Junbo Cui, Hongji Zhu, Tianchi Cai, Haoyu Li, Weilin Zhao, Zhihui He, et~al.
\newblock Minicpm-v: A gpt-4v level mllm on your phone.
\newblock \emph{arXiv preprint arXiv:2408.01800}, 2024.

\bibitem[You et~al.(2024)You, Gu, Li, Cai, Zhu, Dong, and Xue]{you2024descriptive}
Zhiyuan You, Jinjin Gu, Zheyuan Li, Xin Cai, Kaiwen Zhu, Chao Dong, and Tianfan Xue.
\newblock Descriptive image quality assessment in the wild.
\newblock \emph{arXiv preprint arXiv:2405.18842}, 2024.

\bibitem[Yue et~al.(2024{\natexlab{a}})Yue, Ni, Zhang, Zheng, Liu, Zhang, Stevens, Jiang, Ren, Sun, et~al.]{yue2024mmmu}
Xiang Yue, Yuansheng Ni, Kai Zhang, Tianyu Zheng, Ruoqi Liu, Ge~Zhang, Samuel Stevens, Dongfu Jiang, Weiming Ren, Yuxuan Sun, et~al.
\newblock Mmmu: A massive multi-discipline multimodal understanding and reasoning benchmark for expert agi.
\newblock In \emph{Proceedings of the IEEE/CVF Conference on Computer Vision and Pattern Recognition}, pages 9556--9567, 2024{\natexlab{a}}.

\bibitem[Yue et~al.(2024{\natexlab{b}})Yue, Zheng, Ni, Wang, Zhang, Tong, Sun, Yu, Zhang, Sun, et~al.]{yue2024mmmupro}
Xiang Yue, Tianyu Zheng, Yuansheng Ni, Yubo Wang, Kai Zhang, Shengbang Tong, Yuxuan Sun, Botao Yu, Ge~Zhang, Huan Sun, et~al.
\newblock Mmmu-pro: A more robust multi-discipline multimodal understanding benchmark.
\newblock \emph{arXiv preprint arXiv:2409.02813}, 2024{\natexlab{b}}.

\bibitem[Zeng et~al.(2022)Zeng, Liu, Du, Wang, Lai, Ding, Yang, Xu, Zheng, Xia, et~al.]{zeng2022glm}
Aohan Zeng, Xiao Liu, Zhengxiao Du, Zihan Wang, Hanyu Lai, Ming Ding, Zhuoyi Yang, Yifan Xu, Wendi Zheng, Xiao Xia, et~al.
\newblock Glm-130b: An open bilingual pre-trained model.
\newblock \emph{arXiv preprint arXiv:2210.02414}, 2022.

\bibitem[Zhu et~al.(2023)Zhu, Chen, Shen, Li, and Elhoseiny]{zhu2023minigpt}
Deyao Zhu, Jun Chen, Xiaoqian Shen, Xiang Li, and Mohamed Elhoseiny.
\newblock Minigpt-4: Enhancing vision-language understanding with advanced large language models.
\newblock \emph{arXiv preprint arXiv:2304.10592}, 2023.

\bibitem[Zhu et~al.(2025)Zhu, Wang, Chen, Liu, Ye, Gu, Duan, Tian, Su, Shao, et~al.]{zhu2025internvl3}
Jinguo Zhu, Weiyun Wang, Zhe Chen, Zhaoyang Liu, Shenglong Ye, Lixin Gu, Yuchen Duan, Hao Tian, Weijie Su, Jie Shao, et~al.
\newblock Internvl3: Exploring advanced training and test-time recipes for open-source multimodal models.
\newblock \emph{arXiv preprint arXiv:2504.10479}, 2025.

\end{thebibliography}

% \newpage

% \appendix

% \section{Technical Appendices and Supplementary Material}
% Technical appendices with additional results, figures, graphs and proofs may be submitted with the paper submission before the full submission deadline (see above), or as a separate PDF in the ZIP file below before the supplementary material deadline. There is no page limit for the technical appendices.

%%%%%%%%%%%%%%%%%%%%%%%%%%%%%%%%%%%%%%%%%%%%%%%%%%%%%%%%%%%%

% \begin{figure}[t]
% \begin{center}
% \includegraphics[width=0.5\columnwidth]{./images/placeholder.jpg}
% \caption{Examples of questions from different domains in \method{}.}
% \label{fig_motivation}
% \end{center}
% % \vskip -0.2in
% \end{figure}

\newpage

\end{document}